\documentclass{article}
\usepackage[preprint]{corl_2023} 
\usepackage{amsmath,amsfonts}
\usepackage{graphicx}
\usepackage{subcaption}
\usepackage[]{mdframed}
\usepackage[usenames,dvipsnames]{xcolor}
\usepackage[final]{changes}

\author{K. Niranjan Kumar \\
Georgia Institute of Technology \\
\And
Irfan Essa \\
Georgia Institute of Technology \\
\And
Sehoon Ha \\
Georgia Institute of Technology\\
}
\title{Words into Action: Learning Diverse Humanoid Robot Behaviors using Language Guided Iterative Motion Refinement}
\begin{document}



\newcommand{\old}[1]{\textcolor{red}{\textbf {\st{#1}}}}
\newcommand{\new}[1]{\textcolor{black}{#1}}
\newcommand{\note}[1]{\cmt{Note: #1}}
\newcommand{\niranjan}[1]{\textcolor{TealBlue}{#1}}
\newcommand{\sehoon}[1]{\textcolor{orange}{{Sehoon: #1}}}
\newcommand{\sehoonedit}[1]{\textcolor{brown}{{#1}}}
\newcommand{\sbt}[1]{\textcolor{red}{{}}}

\newcommand{\edited}[1]{\textcolor{Darkpink}{{edited: #1}}}
\newcommand{\eqnref}[1]{Equation~(\ref{eqn:#1})}

\long\def\ignorethis#1{}

\newcommand{\etal}{{\em{et~al.}\ }}
\newcommand{\eg}{e.g.\ }
\newcommand{\ie}{i.e.\ }

\newcommand{\figtodo}[1]{\framebox[0.8\columnwidth]{\rule{0pt}{1in}#1}}
\newcommand{\figref}[1]{Figure~\ref{fig:#1}}
\newcommand{\secref}[1]{Section~\ref{sec:#1}}

\newcommand{\vc}[1]{\ensuremath{\boldsymbol{#1}}}
\newcommand{\pd}[2]{\ensuremath{\frac{\partial{#1}}{\partial{#2}}}}
\newcommand{\pdd}[3]{\ensuremath{\frac{\partial^2{#1}}{\partial{#2}\,\partial{#3}}}}
\newcommand{\E}[2]{\operatorname{\mathbb{E}}_{#1}\left[#2\right]}

\newcommand{\vEndEff}{\ensuremath{\vc{d}}}
\newcommand{\vRelMove}{\ensuremath{\vc{r}}}
\newcommand{\sSet}{\ensuremath{S}}

\newcommand{\vControl}{\ensuremath{\vc{u}}}
\newcommand{\vPoint}{\ensuremath{\vc{p}}}
\newcommand{\sSpringCoef}{{\ensuremath{k_{s}}}}
\newcommand{\sDamperCoef}{{\ensuremath{k_{d}}}}
\newcommand{\vHandle}{\ensuremath{\vc{h}}}
\newcommand{\vForce}{\ensuremath{\vc{f}}}

\newcommand{\mTransChain}{\ensuremath{\vc{W}}}
\newcommand{\mRotateTrans}{\ensuremath{\vc{R}}}
\newcommand{\sJoint}{\ensuremath{q}}
\newcommand{\vJoint}{\ensuremath{\vc{q}}}
\newcommand{\mJoint}{\ensuremath{\vc{Q}}}
\newcommand{\mMass}{\ensuremath{\vc{M}}}
\newcommand{\sMass}{\ensuremath{{m}}}
\newcommand{\vGravity}{\ensuremath{\vc{g}}}
\newcommand{\vConstr}{\ensuremath{\vc{C}}}
\newcommand{\sConstr}{\ensuremath{C}}
\newcommand{\vCOM}{\ensuremath{\vc{x}}}
\newcommand{\sGeneralForce}[1]{\ensuremath{Q_{#1}}}
\newcommand{\vStateVar}{\ensuremath{\vc{y}}}
\newcommand{\vControlVar}{\ensuremath{\vc{u}}}
\newcommand{\argmax}{\operatornamewithlimits{argmax}}
\newcommand{\argmin}{\operatornamewithlimits{argmin}}

\newcommand{\tr}[1]{\ensuremath{\mathrm{tr}\left(#1\right)}}
\newcommand{\framedtext}[1]{%
\par%
\noindent\fbox{%
    \parbox{\dimexpr\linewidth-2\fboxsep-2\fboxrule}{#1}%
}%
}
\makeatletter
\let\oldlt\longtable
\let\endoldlt\endlongtable
\def\longtable{\@ifnextchar[\longtable@i \longtable@ii}
\def\longtable@i[#1]{\begin{figure}[t]
\onecolumn
\begin{minipage}{0.5\textwidth}
\oldlt[#1]
}
\def\longtable@ii{\begin{figure}[t]
\onecolumn
\begin{minipage}{0.5\textwidth}
\oldlt
}
\def\endlongtable{\endoldlt
\end{minipage}
\twocolumn
\end{figure}}
\makeatother

\maketitle
\begin{abstract}
Humanoid robots are well suited for human habitats due to their morphological similarity, but developing controllers for them is a challenging task that involves multiple sub-problems, such as control, planning and perception. In this paper, we introduce a method to simplify controller design by enabling users to train and fine-tune robot control policies using natural language commands. We first learn a neural network policy that generates behaviors given a natural language command, such as ``walk forward'', by combining Large Language Models (LLMs), motion retargeting, and motion imitation. Based on the synthesized motion, we iteratively fine-tune by updating the text prompt and querying LLMs to find the best checkpoint associated with the closest motion in history. We validate our approach using a simulated Digit humanoid robot and demonstrate learning of diverse motions, such as walking, hopping, and kicking, without the burden of complex reward engineering. In addition, we show that our iterative refinement enables us to learn $3\times$ times faster than a naive formulation that learns from scratch. See video of results here - \href{https://www.kniranjankumar.com/words_into_action/}{https://www.kniranjankumar.com/words\_into\_action/}
\end{abstract}
\section{INTRODUCTION}
Humanoid robots have long fascinated both scientists and science fiction writers. Recent advancements have led to controllers that enable these robots to execute complex maneuvers like jumping, hopping, and even front-flips. However, the traditional approach to designing these controllers involves labor-intensive engineering: creating a robot model, planning trajectories, and optimizing a cost function. This process must be repeated for each new motion, making it an unscalable solution for controllers that need to adapt to changing environments and skill sets~\cite{posa2016optimization, henze2015approach}.
An alternative is to use model-free deep Reinforcement Learning (DRL) techniques that learn control policies by maximizing a reward function through extensive simulation data. While these techniques have shown promise in solving complex control problems, they come with their own challenges, particularly in reward engineering. Designing a reward function that captures the nuances of desired behaviors for high-DOF robots is no small feat. 

We take inspiration from the recent advances in the control of virtual humanoids by imitating the given reference motion~\cite{peng2021amp,peng2018deepmimic}, which we refer to as motion imitation. This approach enables efficient learning of a variety of motor skills by offering a unified task definition of imitating the corresponding motions.
Recent studies have explored using motion capture data as a reference for learned policies, showing promising results in quadruped robot locomotion~\cite{escontrela2022adversarial,klipfel2023learning}.

Building on these insights, our work focuses on learning joint-level control policies for humanoid robots directly from language commands, significantly simplifying the reward engineering process. Our method allows for iterative fine-tuning of policies through interactive language commands, offering precise control over the robot's behaviors. We achieve this by first generating human motion based on language commands, retargeting them to a humanoid robot, and then learning control policies using a motion imitation approach.
We then develop language-based policy refinement, a process which allows users to adjust the learned behavior by identifying the closest checkpoint with a large language model and fine-tune the policy starting from it.

We demonstrate a wide range of behaviors for the humanoid robot Digit using our framework. From simple language prompts, our framework learns diverse behaviors such as hopping, stepping to the side, and kicking.
In addition, we provide evidence supporting the benefits of iterative human-guided policy refinement approach, which offers $3\times$ better sample efficiency.

\section{RELATED WORK}
\subsection{Robot learning for legged robots}
A natural solution to building robots adapted to the human living environment, is to adopt a morphology similar to human form. 
However, designing controllers for humanoid robots that are versatile and robust to external perturbations is a challenging problem that has been studied extensively over the last few decades. A classical approach is to develop a dynamics model for the robot and then use it to develop controllers that plan and execute control actions, optimizing a specified objective. These models~\cite{grizzle2014models,hurmuzlu2004modeling} span from simple approximations that reduce the robot to a linear inverted pendulum~\cite{raibert1986legged,kajita2003biped, yamaguchi1999development, sugihara2002real,hirai1998development} to more sophisticated alternatives that consider the entire dynamics of the humanoid~\cite{posa2016optimization, henze2015approach} or some combination of the two strategies~\cite{dai2014whole}. While simpler models make the problem tractable from a computational standpoint, they sacrifice exploiting the full capabilities of the robot dynamics and constrain themselves to motion feasible on the simple model. Modeling the full dynamics of the robot, on the other hand, is difficult and time-consuming and may not be applicable for real-time robot control. 

In recent years, learning-based approaches, particularly deep reinforcement learning (DRL)~\cite{sutton2018reinforcement} have gained increasing attention from the humanoid robot control community, following impressive results on quadruped~\cite{tan2018sim,lee2020learning,kumar2023cascaded} and bipedal robots~\cite{li2021reinforcement,kumar2022adapting}.
Some work incorporates learning within other model-based frameworks to improve robustness and adaptability. For example, in~\cite{castillo2022reinforcement} a learned policy generates actuator trajectories that are then tracked using a feedback regulator designed with the robot model. In a related work~\cite{krishna2022linear}, a learned policy acts as a foot-trajectory modulator while a low-level gait controller regulates the torso and ankle orientation.
A recent line of work takes an end-to-end learning-based framework to develop humanoid robot controllers~\cite{radosavovic2023learning,guo2023decentralized}. Unlike model-based approaches, learning based techniques do not require solving for the control output at every time-step using the robot dynamics model, which makes them suitable for high-frequency real-time control. These policies can be trained in massively parallelized simulations~\cite{makoviychuk2021isaac} and transferred directly to the real-world without additional fine-tuning. 
While learning-based approaches offer a powerful alternative to model-based control, it is often challenging to design and tune reward functions to accomplish a given task. Recent work,~\cite{escontrela2022adversarial,peng2020learning} leverages expert demonstrations to ground the space of behaviors that emerge from the policy by imposing a motion prior.
\subsection{Human motion generation}
Generating human motion has a wide range of applications from animation~\cite{lee2002interactive} to modeling human behaviors for human robot interaction. With the abundance of human motion capture datasets like HumanML3D~\cite{guo2022generating} researchers have been able to create generative model for human motion~\cite{zhang2023generating}. With the advent of Large Language Models (LLMs), using text to generate human motion has gained considerable interest within the research community. Given the ease of prompting and guiding the generation process, LLMs offer a promising strategy to control synthesis.

However, the generated motion from these methods is not grounded in physics and often demonstrates foot sliding, jitters or self-collisions. So a line of work focuses on learning control policies for virtual agents in a physics simulation to imitate motion capture trajectories while being physically plausible~\cite{peng2021amp,peng2018deepmimic}. In this work, we build upon these ideas to use human motion as a reference to guide on a humanoid robot control policies.

\subsection{LLMs for Robot Control}
LLMs have demonstrated human-level reasoning capabilities across a wide range of domains ranging from code-generation~\cite{chen2021evaluating}, multi-step reasoning~\cite{wei2022chain,kojima2022large,singh2023progprompt} to prompt-engineering~\cite{zhou2022large}. One limitation of LLMs is the lack of grounding in the real world. Robotics offers a promising approach to ground LLMs into the real world. They have been rapidly adopted by the robotics research community in the past few months. With LLMs reaching human-level reasoning, they have the potential to significantly augment a robot's capabilities. They have been shown to generate long-horizon plans~\cite{ahn2022can}, understand and process complex cluttered scenes~\cite{driess2023palm}, and explain the reasoning behind decisions a robot takes~\cite{zhang2023explaining}. In this work, we use LLMs to generate human motion trajectories to build prompts based on interactive human instructions and to guide model initialization to reduce training time.
\section{LEARNING POLICIES FROM LANGUAGE PROMPTS}
The first step is to learn control policies from the language prompts given by users. It consists of three main components:
\begin{figure}
    \centering
    \includegraphics[width=0.99\linewidth]{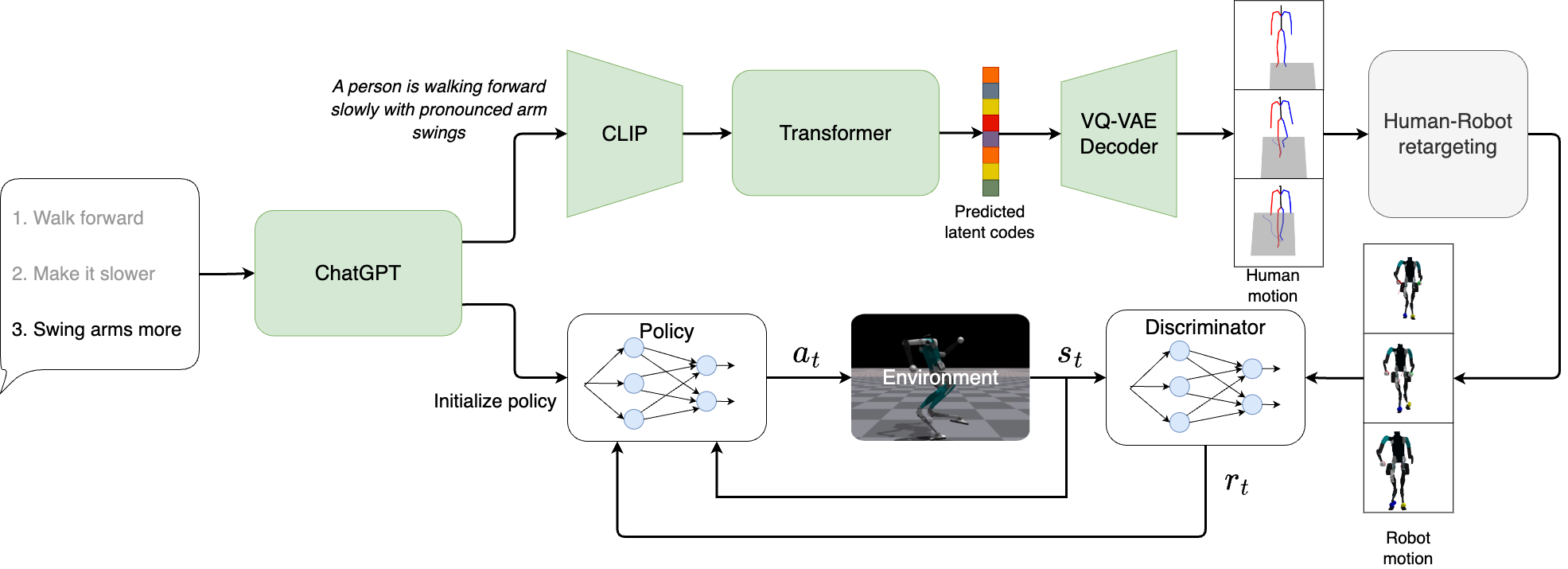}
    \caption{Overview of our proposed approach. Given a language instruction our framework outputs a learned control policy for the corresponding behavior.}
    \label{fig:architecture}
\end{figure}
\begin{enumerate}
    \item Human motion generation from text input
    \item Motion retargeting from human to robot
    \item Training a control policy to imitate retargeted motion
\end{enumerate}
We provide an overview of our framework in Fig.~\ref{fig:architecture}. In the following section, we discuss each of these components.
\subsection{Human motion generation from text input}
Human motion generation has a wide range of applications from animation to behavior modeling. A natural modality to direct the generation of motion is through textual descriptions. Recently, there has been rapid progress in generating diverse high-quality human motion trajectories from language descriptions. Most of these approaches leverage large-scale annotated datasets of motion-capture data to train generative models to map language input to a distribution of possible human trajectories.

In this work, we follow T2M-GPT~\cite{zhang2023generating} to generate human trajectories from a language description. We first train a VQ-VAE that builds a discrete latent representation of the human motion space. By selecting different parameters in the latent space, we have a handle on the nature of the motion generated. For a motion sequence $X=[x_1,x_2,\cdots,x_T]$ where $x_t\in\mathbb{R}^d$, $T$ is the number of frames in the motion sequence, $d$ is the dimension of the motion, the goal is to learn a codebook $C=\{c_k\}^K_{k=1}$ of size $K$ with $c_k\in\mathbb{R}^{d_c}$ where $d_c$ is the dimension of the codes. To accomplish this, we train an autoencoder type architecture where an encoder $E$ computes the latent variable $Z=E(X)$ from $X$ and a decoder $D$ reconstructs the motion from the latent variable. In contrast to a traditional variational autoencoder, the latent variable is discrete, defined as $Z=[z_1,z_2,\cdots,z_{T/l}]$ where $z_i\in\mathbb{R}^{d_c}$ and $l$ is the temporal downsampling rate. The encoded $Z$ is then quantized to $\hat{Z}$ by finding the closest element in $C$.
\begin{equation}
\hat{z_i}=\underset{c_k \in C}{\arg \min }\left\|z_i-c_k\right\|_2
\end{equation}
The VQ-VAE objective is then to minimize the following:

\begin{equation}
            \mathcal{L} = \mathcal{L}_{\textsubscript{\it{re}}} + \mathcal{L}_{\textsubscript{\it{embed}}} + \beta \mathcal{L}_{\textsubscript{\it{commit}}}\\
\end{equation}  
where,
\begin{equation}
    \begin{aligned}
    \mathcal{L}_{\textsubscript{\it{embed}}} & = \left\|Z-sg[\hat{Z}]\right\|_2 ,
    \mathcal{L}_{\textsubscript{\it{commit}}} = \left\|sg[Z]-\hat{Z}\right\|_2
    \end{aligned}
\end{equation}
$\mathcal{L}_{\textsubscript{\it{embed}}}$ minimizes the embedding loss so that the predicted $Z$ is close to the elements in the codebook and $\mathcal{L}_{\textsubscript{\it{commit}}}$ updates the codebook to better fit the encoded values. $\beta$ is a hyper-parameter to control the relative impact of these terms on the final loss. The reconstruction loss $\mathcal{L}_{\textsubscript{\it{re}}}$ is an $L_1$ smooth loss between the ground truth and the predicted position and velocity of the motion.

Given an expressive enough motion embedding, we can generate any arbitrary motion sequence by auto-regressively generating a series of codebook indices. 
Including a text condition $c$ provides a handle on the generated codebook entries. This process is modeled using transformer architecture and trained on the HumanML3D dataset~\cite{guo2022generating}.
Please refer to T2M-GPT~\cite{zhang2023generating} for more details about training and architecture.
\subsection{Motion retargeting from human to robot}
Given a human motion trajectory generated from the previous step, we want to imitate it using a learned control policy. Due to differences in the skeletons, we cannot directly map the joint angles. We instead track just the end-effector locations i.e., $2$ hands + $2$ legs. 
 We design an Inverse Kinematics (IK) objective that minimizes error between the reference relative end-effector positions and the robot end-effector positions. Given the list of reference positions $x_{\text{H}}$ and robot joint angles $q_{\text{D}}$ the IK objective is defined as follows:
\begin{equation}
    q^* = \underset{q}{\arg \min} \left\| x_H - T_{\text{FK}}(q)\right\|_2 + \lambda \mathcal{C}_{\text{f}}(q)
\end{equation}
$T_{\text{FK}}(q)$ is the forward kinematics function of the robot, $C_{\text{f}}$ is a term that captures the feasibility of a pose and $\lambda$ is a scaling factor.
$C_\text{f}$ in our case simulates rod constraints to mimic the 4-bar linkage present in the Digit robot and ensures that change in $q$ over consecutive timesteps in minimal.

We solve this optimization for every timestep in the trajectory using Sequential Least Squares Programming. To ensure the temporal smoothness, we initialize the variables with the solution of the previous timestep. While the trajectory we get at the end is kinematically feasible, we cannot guarantee its success when subjected to robot dynamics and physics. Hence, we train a neural network policy to track these reference poses while being embedded in a physics simulator.
In the next section, we discuss our policy learning framework.

\subsection{Training control policy to imitate retargeted motion}
Given a reference trajectory of the robot, we want to train a control policy that imitates it while being dynamically feasible. It should be noted that some of these states might not be physically realistic on the robot. We would like our approach to ignore these states and focus on parts of the trajectory that can be reliably tracked instead. Motion imitation approaches offer a promising solution to this problem. Rather than forcing the policy to reach every segment in the reference, motion imitation approaches typically only require that the trajectories generated from the policy resemble those found in the reference. Therefore, the infeasible segments can be ignored as the policy is trained. Thus, the \textit{mode collapse} problem that often plagues generative models works to our advantage here, eliminating the need  to manually curate a set of dynamically feasible motions for the robot.

We use Adversarial Motion Priors (AMP)~\cite{peng2021amp} to train neural network policies for joint-level control of the robot. Given a dataset of state transitions $\mathcal{M}$,  a discriminator network $D(s,s^\prime)$ is trained to predict if the transition is from the learned policy $\pi$ or from $\mathcal{M}$. This follows an architecture similar to GAIL~\cite{ho2016generative} but with state transitions $(s,s^\prime)$ instead of state-action pairs $(s,a)$. 
\begin{equation}
\underset{D}{\arg \min }-\mathbb{E}_{d^{\mathcal{M}}\left(\mathrm{s}, \mathrm{s}^{\prime}\right)}\left[\log \left(D\left(\mathrm{~s}, \mathrm{~s}^{\prime}\right)\right)\right]-\mathbb{E}_{d^\pi\left(\mathrm{s}, \mathrm{s}^{\prime}\right)}\left[\log \left(1-D\left(\mathrm{~s}, \mathrm{~s}^{\prime}\right)\right)\right] .
\end{equation}
The discriminator is then used to compute reward during policy training. 
\begin{equation}
\label{ed:reward}
r\left(s_t, s_{t+1}\right)=\max [0, 1-0.25(D(s_t,s_{t+1})-1)^2]
\end{equation}
The neural network policy is learned with proximal-policy optimization(PPO)~\cite{schulman2017proximal} using a value function network and a gradient penalty to stabilize training. See AMP~\cite{peng2021amp} for network architecture and details about the training process.

\section{LANGUAGE-GUIDED ITERATIVE POLICY REFINEMENT}
Training control policies for high-DOF robots involves careful reward engineering to tune the style of the behaviors that emerge, requiring expert knowledge about the impact of different reward terms on the final policy. 
In this section, we seek to simplify this process by conditioning the behaviors generated on natural language commands, iteratively adjusting the trained policy. We leverage the general-purpose language understanding and reasoning capacity of LLMs to achieve this goal. Given an input command update, the LLM is assigned two tasks: 
\begin{enumerate}
    \item Create/update the instruction given to T2M-GPT to incorporate the update
    \item Initialize the policy with the closest previously trained model (if applicable) to improve sample efficiency.
\end{enumerate}
We prompt a conversational LLM (ChatGPT-4) to behave as a prompt generator and policy initalizer using the following prompt: 
\framedtext{
\textit{Your role is to generate prompts for a motion model. You have to continuously create/update the original prompt based on user input ``command" and recover a motion ``closest\_prompt\_in\_history" from a history of generated prompts ``motion\_history" that resemble the generated prompt. If the motions in the history are completely unrelated to the generated prompt, return ``None". Your prompts describe the actions of a person. Examples of prompts:}

\begin{enumerate}
    \item \textit{A person is walking forward}
    \item \textit{A person is hopping forward then turning around and hopping back to the start.}
\end{enumerate}
\textit{The user commands will have the format: }
$\langle \textit{command}\rangle$

\textit{You should return :}
\begin{itemize}
    \item $\langle \textit{prompt}\rangle$
    \item $\langle \textit{closest\_prompt\_in\_history}\rangle$
    \item $\langle \textit{motion\_history}\rangle$
\end{itemize}
\textit{``motion\_history"is a list of the motion history that includes the generated``prompt". 
``closest\_prompt\_in\_history" can be ``None" if none of the motions are similar enough to the prompt. For example, jumping is very different from walking. But walking slow is similar to walking fast.Wait for the next user input.}
}

Given new instructions conversational LLM automatically updates the prompt given to T2M-GPT and returns the closest model to initialize the policy with, if it exists. We present the behaviors trained using our approach in the following section.

\section{EXPERIMENTS}
In our work, we train control policies for the humanoid robot Digit from Agility robotics. Digit has ``bird-legs" (digitigrades) which differ significantly from human legs. The robot walks on its toes instead of its foot. We show that our approach is immune to such differences in morphology and can learn a diverse set of joint-level control policies for Digit. We first define the underlying Markov Decision Process (MDP) for our control policy, then describe the simulation setup and results.
\begin{figure}[t]
\centering
    \begin{subfigure}[h]{0.24\linewidth}
    \includegraphics[width=0.99\linewidth]{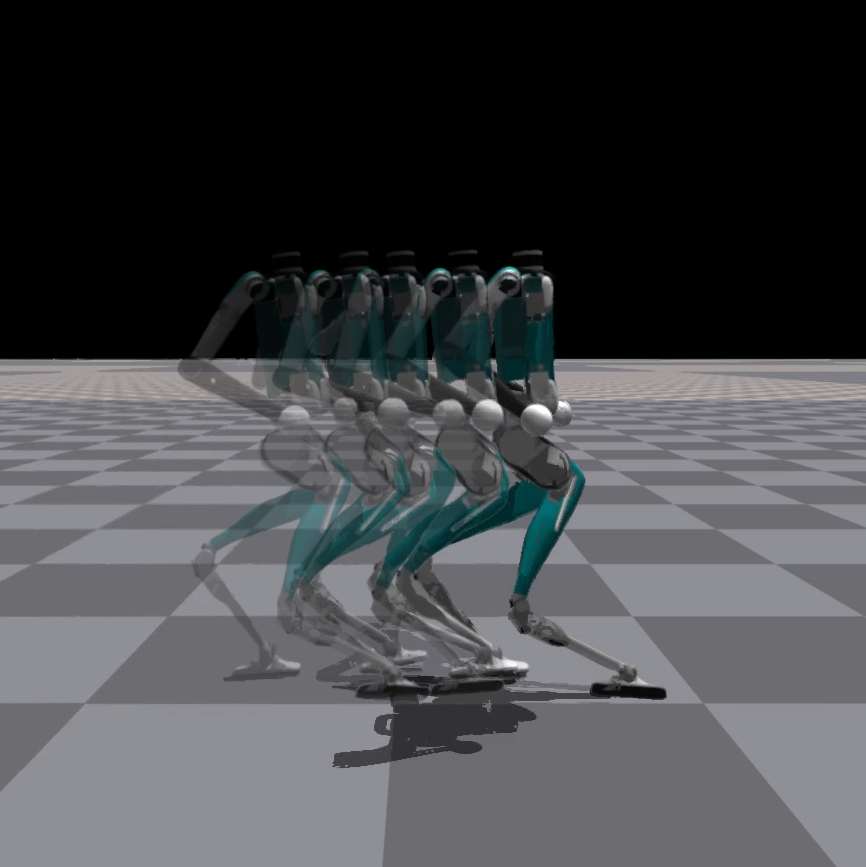}
    \caption{Walk forward}
    \end{subfigure}
    \begin{subfigure}[h]{0.24\linewidth}
    \includegraphics[width=0.99\linewidth]{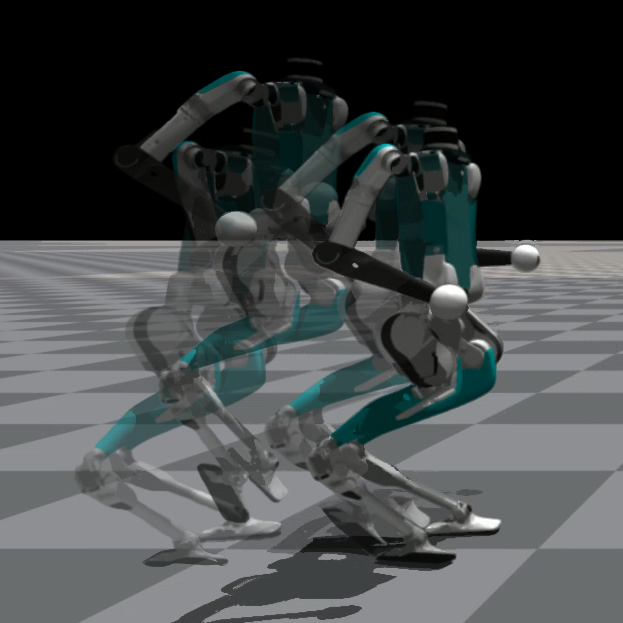}
    \caption{Hop forward}
    \end{subfigure}
    \begin{subfigure}[h]{0.24\linewidth}
    \includegraphics[width=0.99\linewidth]{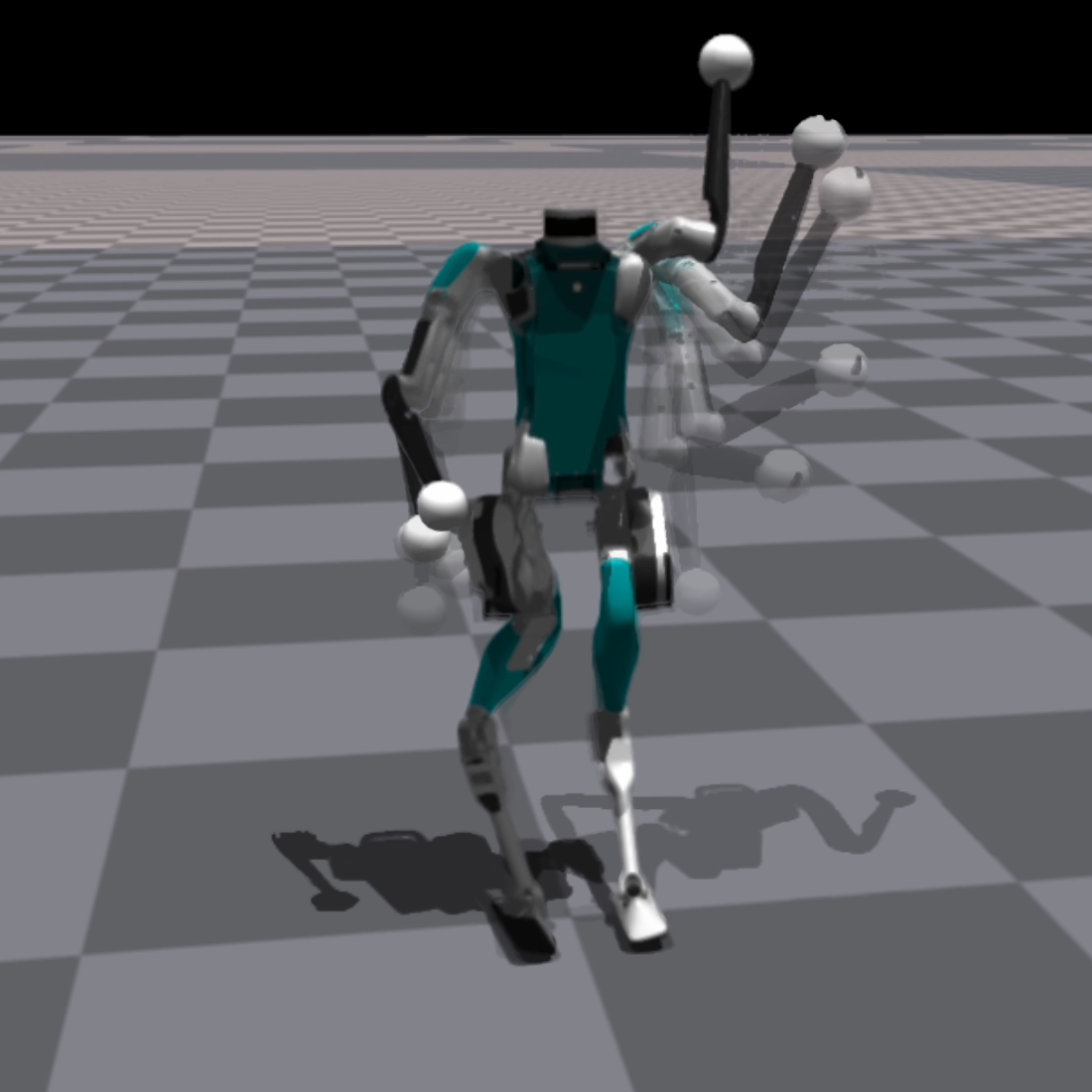}
    \caption{Raise hand}
    \label{fig:raise_hand}
    \end{subfigure}
    \begin{subfigure}[h]{0.24\linewidth}
    \includegraphics[width=0.99\linewidth]{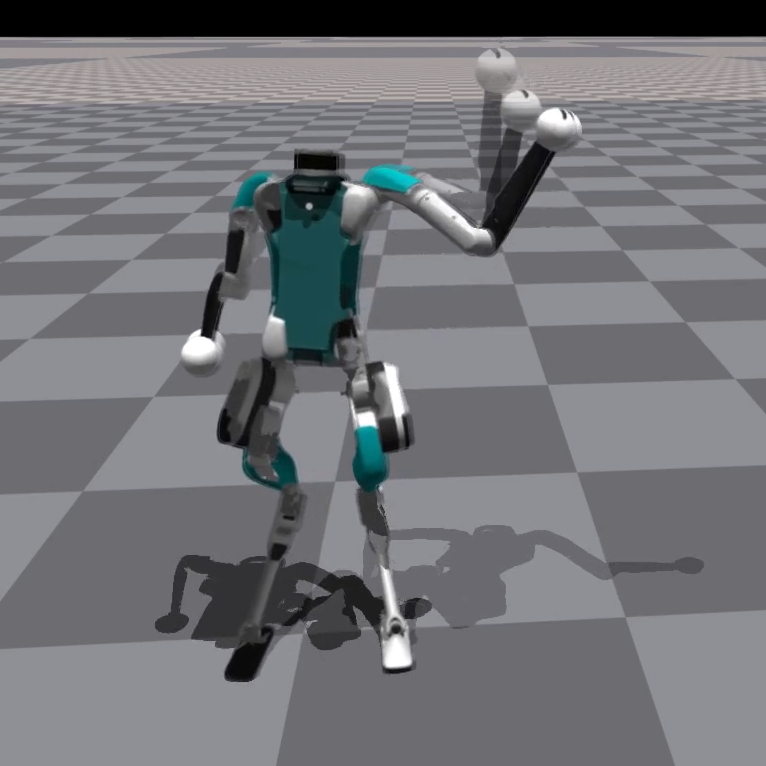}
    \caption{Wave to a friend}
    \label{fig:wave}
    \end{subfigure}
    \centering
    \begin{subfigure}[h]{0.32\linewidth}
    \includegraphics[width=0.99\linewidth]{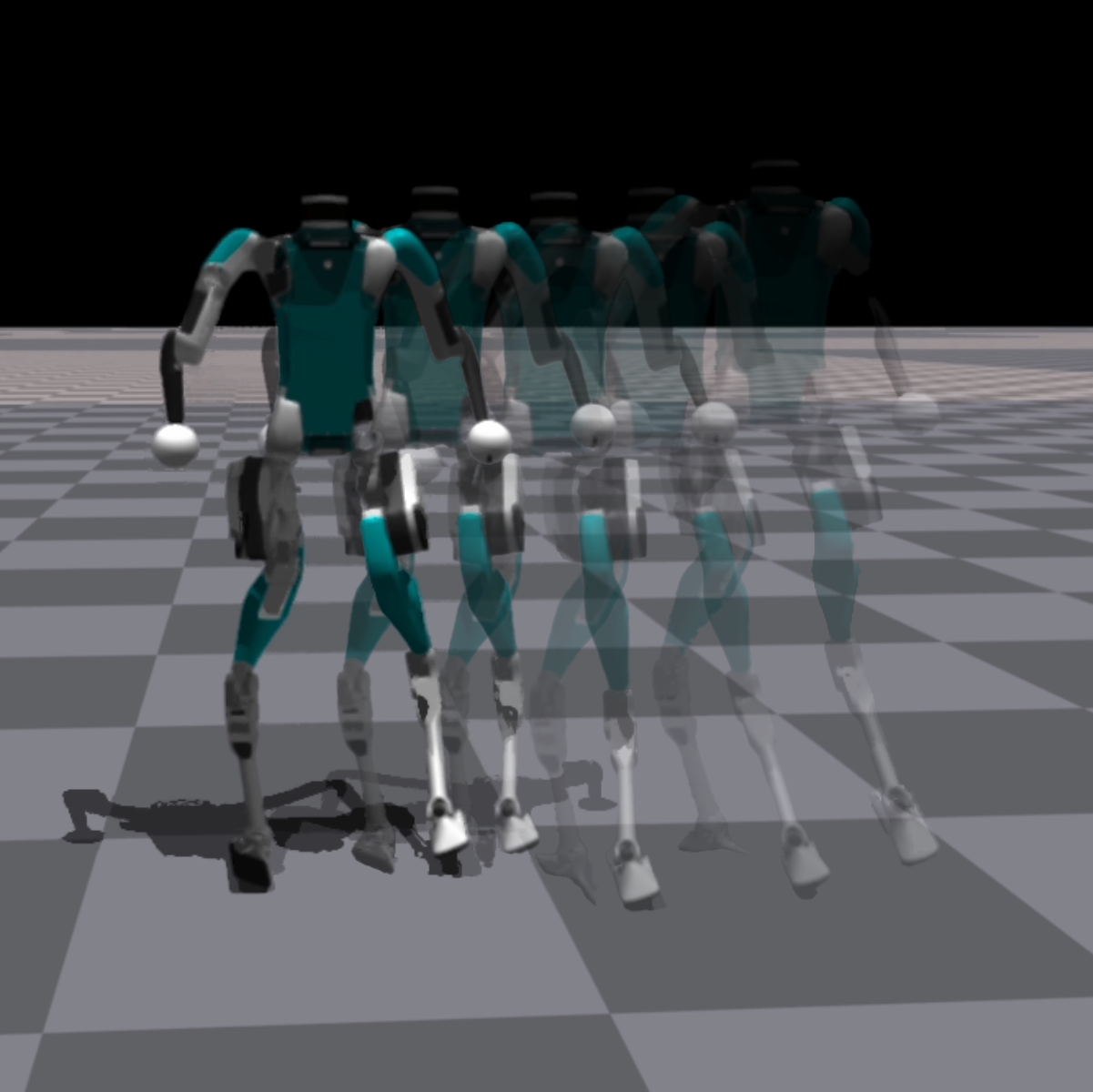}
    \caption{Step to the side gradually}
    \end{subfigure}
    \begin{subfigure}[h]{0.32\linewidth}
    \includegraphics[width=0.99\linewidth]{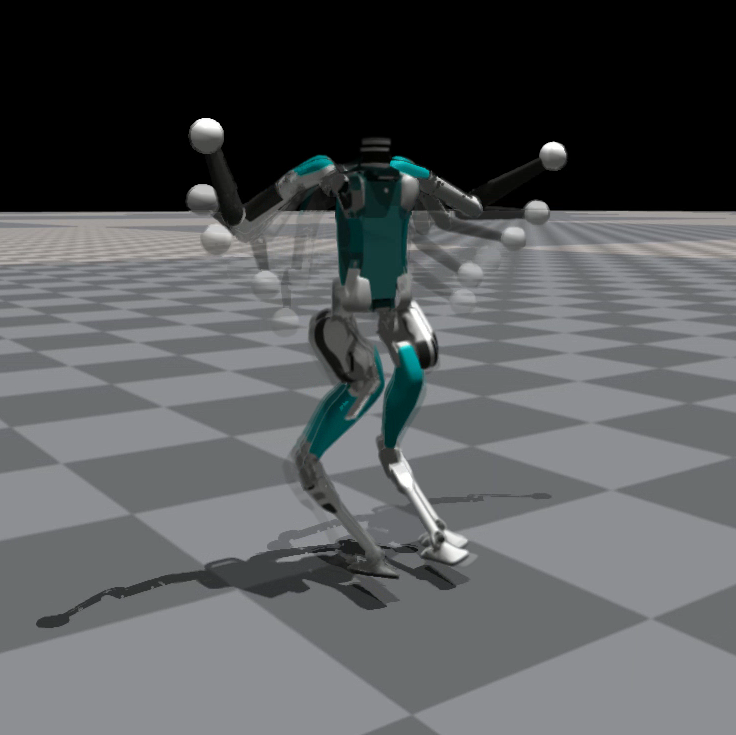}
    \caption{Celebrate}
    \end{subfigure}
    \begin{subfigure}[h]{0.32\linewidth}
    \includegraphics[width=0.99\linewidth]{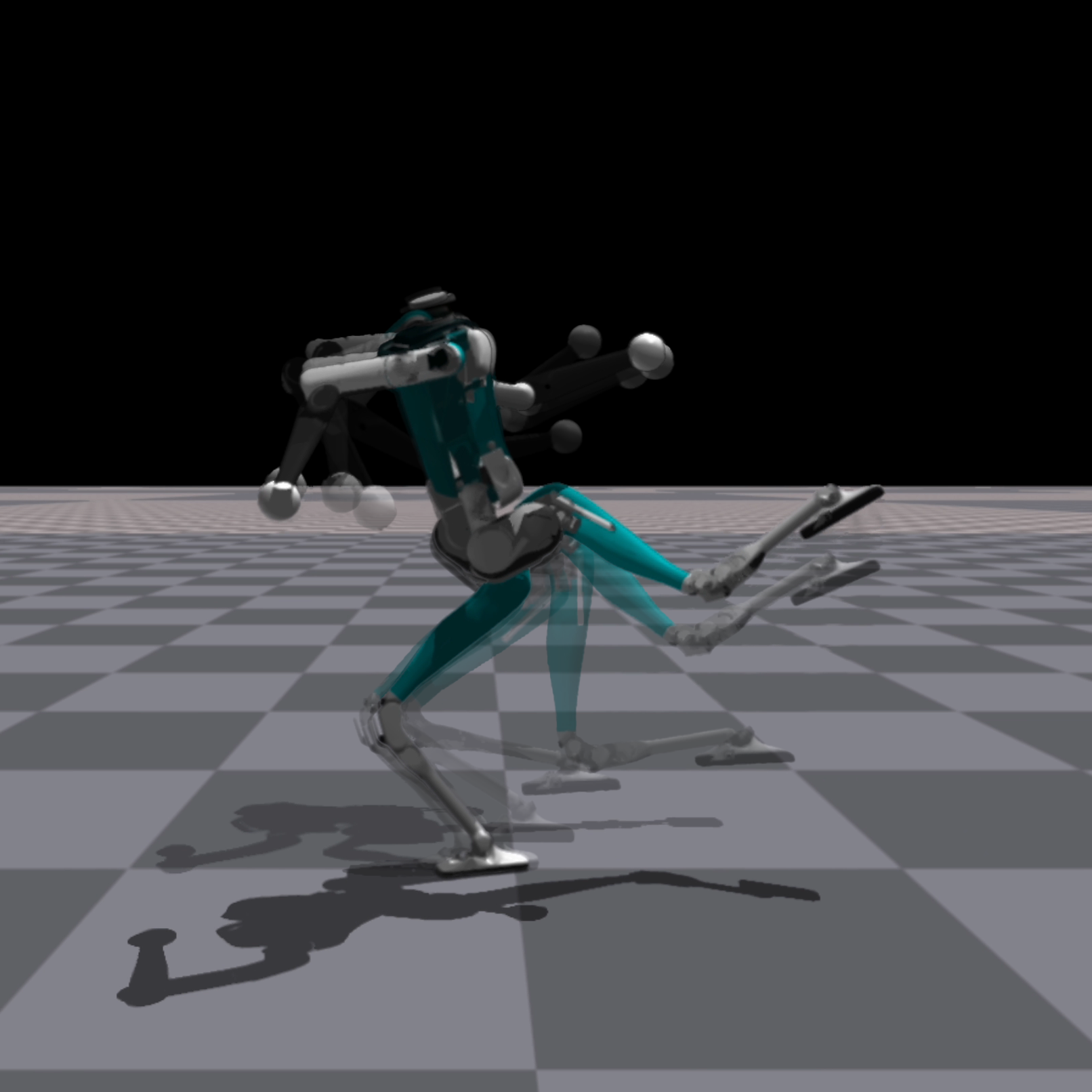}
    \caption{Kick slowly}
    \label{fig:kick}
    \end{subfigure}
    \caption{Motion frames demonstrating the skills learned with our approach}
    \label{fig:results_motion_frame}
\end{figure}
\subsection{Problem Formulation}
Learning control policies for a robot can be modeled as a Markov Decision Process (MDP) where the actions taken by the robot moves it from one state to another. In this work, we define the MDP underlying the robot control problem as follows:
\begin{itemize}
    \item \textbf{States:} The state is a $69$ dimensional vector formed by the concatenation of the root height ($1$), root orientation in normal-tangent encoding ($6$), root velocity ($3$), root angular velocity ($3$), joint positions ($22$), joint velocities ($22$) and end-effector positions ($4\times3$).
    \item \textbf{Actions:} The action is a $22$ dimensional vector of joint angle positions that are sent to a PD controller to calculate the control torque.
    \item \textbf{Rewards:} We use a single reward term calculated using Eq.~\ref{ed:reward}. However, more task specific rewards can be added if required.
\end{itemize}

\subsection{Simulation}
We use IsaacGym~\cite{makoviychuk2021isaac} to train all our control policies on multiple parallelized simulated robots. We simulate $4096$ Digit robots in parallel to train our neural networks on a single NVIDIA Titan X GPU. 
However, IsaacGym does not support closed-loop chains which are necessary to model the $4$-bar linkages in Digit's legs. To handle this, we simulate virtual springs that mimic the physics of the rods, similar to previous work~\cite{radosavovic2023learning}. To ensure that the policies learned are robust, we perform domain randomization by adding Gaussian noise to observation, gravity, and actions taken by the robot with standard deviations $0.02, 0.4, 0.02$, respectively. 
We train each of our policies for a total of $330M$ steps, taking $\approx3$ hours of training time.

\subsection{Results}
\subsubsection{Learned Skills}
Using our framework, we train a diverse set of skills for Digit. We list the behaviors generated using our approach below:
\begin{enumerate}
    \item Walk forward -  a locomotion policy that makes the robot walk forward while swinging its arms
    \item Hop forward -  a policy for that makes the robot hop forward continuously
    \item Raise your hand - a policy that makes the robot stand in place and balance while lifting its hand
    \item Wave to a friend - a policy that makes the robot wave its hand to say hello
    \item Step to the side gradually - a policy that makes the robot step to the side orthogonal to its heading direction.
    \item Celebrate - a policy that makes the robot act like a cheerleader
    \item Kick slowly - a policy that makes the robot balance on one leg as it lifts the other leg up and kicks
\end{enumerate}
\begin{figure}
    \centering
    \begin{subfigure}[h]{0.45\linewidth}
    \includegraphics[width=0.99\linewidth]{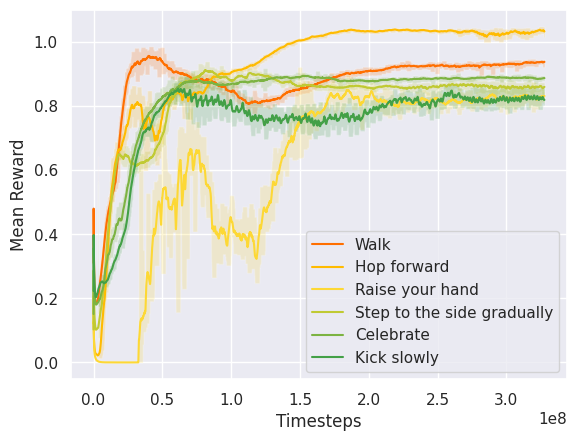}
    \caption{Reward curves for the different skills in our approach}
    \label{fig:rewards}
    \end{subfigure}
    \begin{subfigure}[h]{0.45\linewidth}
    \includegraphics[width=0.99\linewidth]{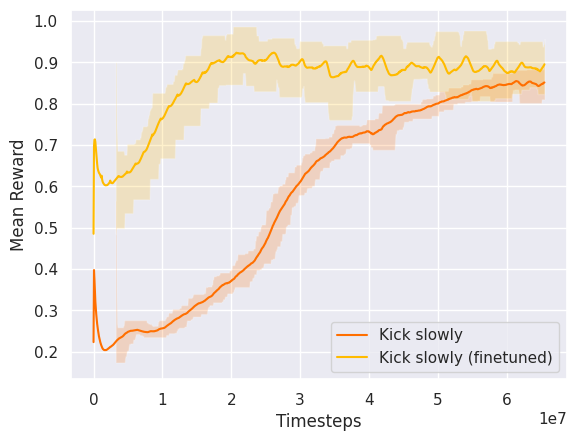}
    \caption{Iteratively fine-tuning a skill from similar previous skills reduces training time}
    \label{fig:iterative_training}
    \end{subfigure}
    \caption{Reward curves for the different behaviors trained using our approach}
\end{figure}
We present motion frames of these skills, along with the intermediate human motion generated in Fig.~\ref{fig:results_motion_frame}. The reward curve for all our policies converge resulting in motions that recreate feasible portions of the reference (See Fig.~\ref{fig:rewards}). Our approach can handle abstract commands and generate meaningful policies that demonstrate the intent behind the command. When instructed to ``Celebrate", our framework generates a cheerleader's routine, hopping and swinging its arms as shown in Fig.~\ref{fig:results_motion_frame}.

\subsubsection{Human guided iterative refinement}
\begin{figure}
    \begin{subfigure}[h]{0.5\linewidth}
    \includegraphics[width=0.96\linewidth]{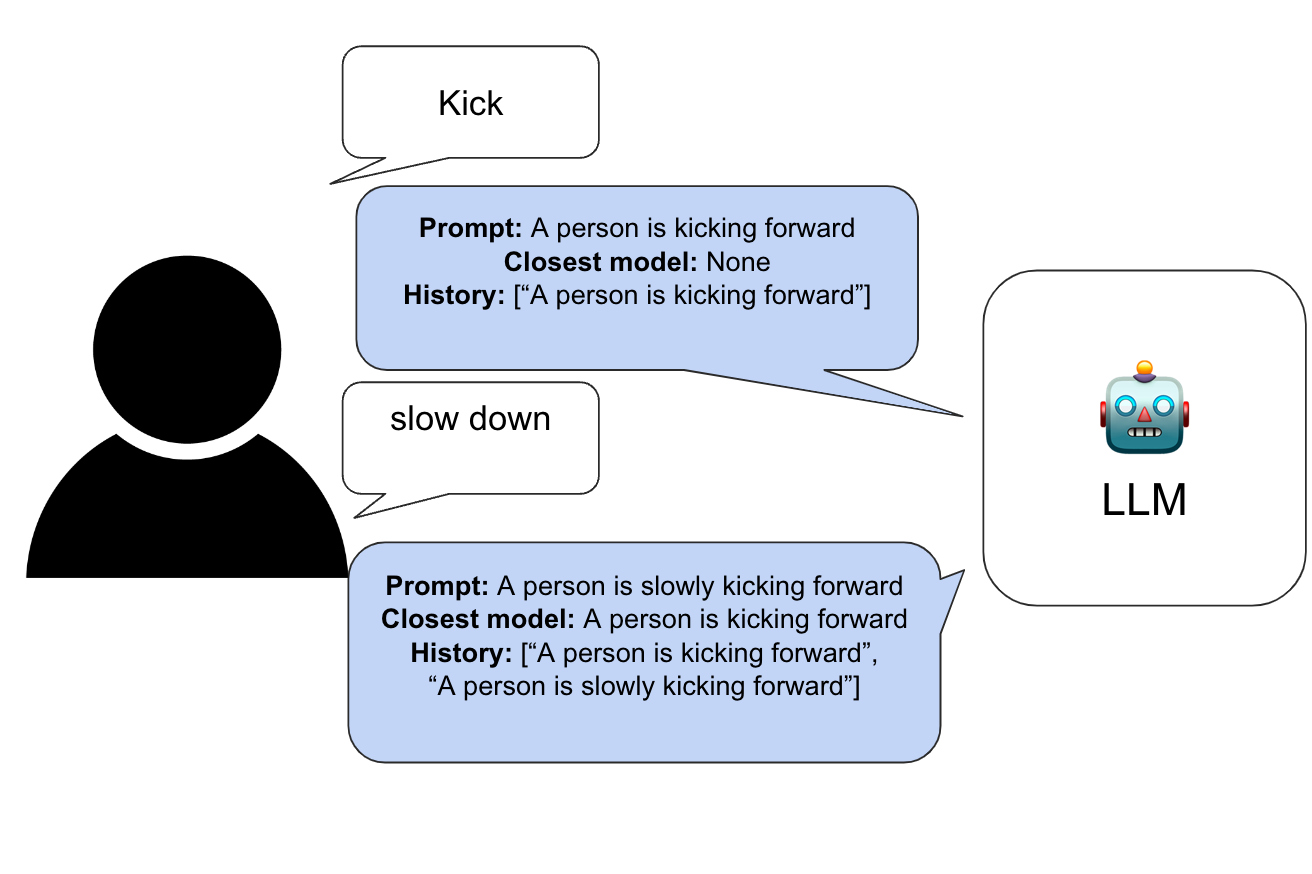}
    \caption{Iterative refinement to generate feasible behavior}
    \label{fig:iterative_refinement_commands}
    \end{subfigure}
    \begin{subfigure}[h]{0.5\linewidth}
    \includegraphics[width=0.96\linewidth]{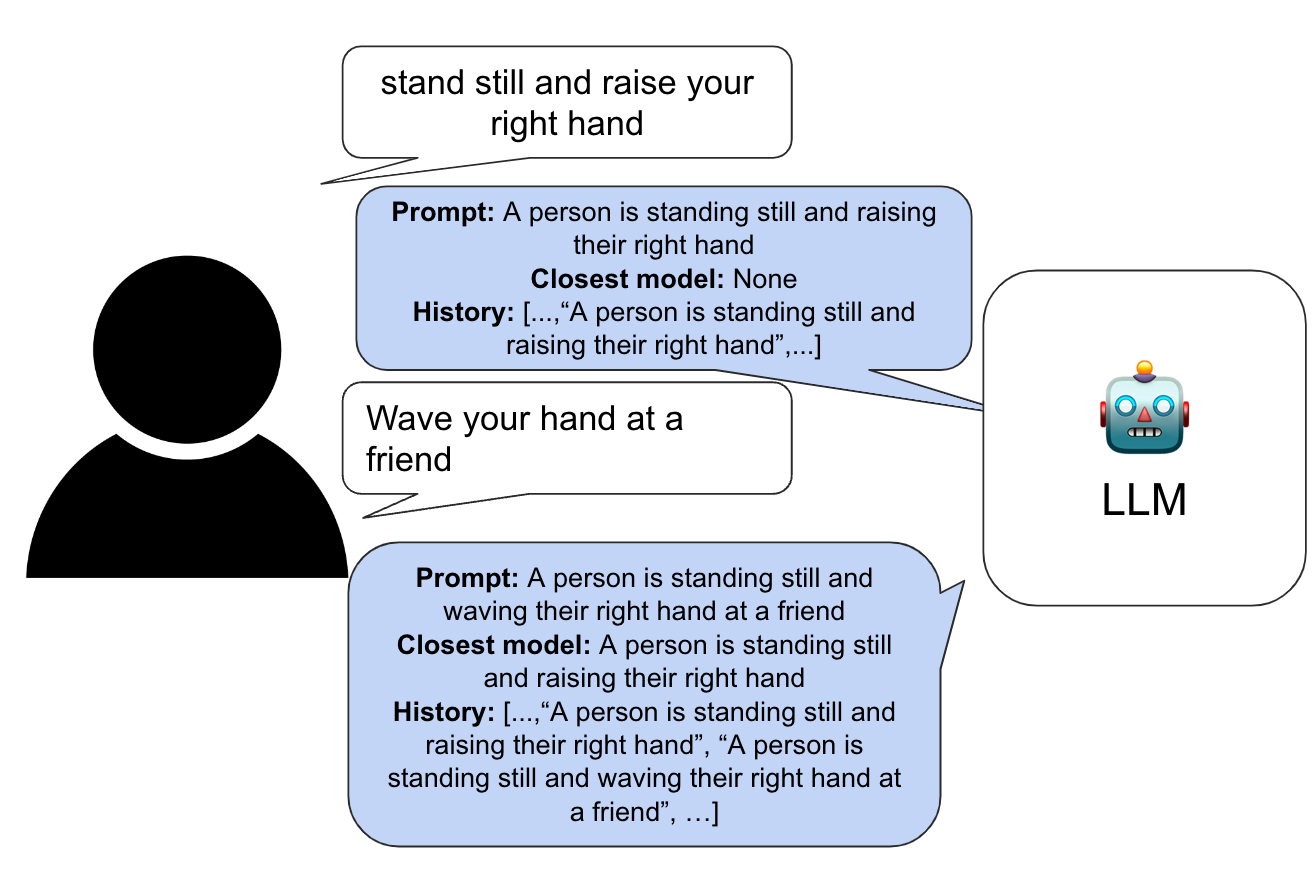}
    \caption{Iterative refinement to reuse learned skills}
    \label{fig:iterative_refinement_commands}
    \end{subfigure}
    \caption{Examples of human guided policy refinement using our framework}
\end{figure}
In addition to generating behaviors given a language instruction, our framework allows us to interactively fine-tune the robot motion.
We show examples of interactive refinement in Fig.~\ref{fig:iterative_refinement_commands}. Notice that as we interact with our framework, the generated command is continuously updated to account for user feedback and the policy is initialized automatically using previously learned behaviors to minimize training iterations. We demonstrate two examples of iterative skill refinement below:
\begin{enumerate}
    \item \added{We tell our framework to make the robot ``Kick”. The LLM does not have any previously learned skill it can reuse, so it initializes the model from scratch. But as the policy trains we notice that the behavior is too sudden and is not feasible on the robot morphology. We tell our framework to ``Slow down”. The LLM updates the command prompt given to T2M-GPT to kick slowly, realizes that we have already learned how to kick and initializes the policy with pre-trained weights. A motion frame for the trained skill is shown in Fig.~\ref{fig:kick}}
     \item \added{We tell our framework to make the robot ``Wave to a friend" after learning a set of skills. Our framework realizes that we have already learned how to ``Stand still and raise your right hand up" (Fig.~\ref{fig:raise_hand}), and initializes the policy with pre-trained weights. A motion frame for the trained skill is shown in Fig.~\ref{fig:wave}}
\end{enumerate}
In Fig.~\ref{fig:iterative_training} we show the reward curves for training a ``Kick slowly" skill from scratch, vs fine-tuning from a previously learned ``Kick" skill. Training using our approach takes significantly fewer steps ($\approx40$M fewer in this case) compared to training from scratch. 

\section{CONCLUSION}
In this work, we presented preliminary results on a framework that generates control policies for dynamic and agile behaviors on a humanoid robot from language instruction. We demonstrated the behaviors learned by our approach on $6$ different instructions and discussed an iterative human-in-the-loop refinement approach to fine-tune the learned behaviors. For future work, we are interested in building a robot motion embedding that directly translates language commands into control actions, without requiring us to retrain a new policy for every instruction. We would also like to deploy our policy on a real Digit robot to test the robustness of our learned policy against environmental perturbations and the sim-to-real gap.
\bibliography{main}  

\end{document}